\documentclass{sig-alternate-05-2015}

\begin{document}


\title{Towards Semantic Integration of Heterogeneous Sensor Data with Indigenous Knowledge for Drought Forecasting}

%
%
%
%
%

\numberofauthors{2} 
%
\author{
%
%
\alignauthor
Adeyinka K. Akanbi\\
       \affaddr{Department of Information Technology}\\
       \affaddr{Central University of Technology}\\
       \affaddr{Free State, South Africa}\\
       \email{aakanbi@cut.ac.za}
\alignauthor
Muthoni Masinde\\
       \affaddr{Department of Information Technology}\\
       \affaddr{Central University of Technology}\\
       \affaddr{Free State, South Africa}\\
       \email{emasinde@cut.ac.za}
}
\maketitle
\begin{abstract}
In the Internet of Things (IoT) domain, various heterogeneous ubiquitous devices would be able to connect and communicate with each other seamlessly, irrespective of the domain. Semantic representation of data through detailed standardized annotation has shown to improve the integration of the interconnected heterogeneous devices. However, the semantic representation of these heterogeneous data sources for environmental monitoring systems is not yet well supported. To achieve the maximum benefits of IoT for drought forecasting, a dedicated semantic middleware solution is required. This research proposes a middleware  that semantically represents and integrates heterogeneous data sources with indigenous knowledge based on a unified ontology for an accurate IoT-based drought early warning system (DEWS). 

\end{abstract}

%
%


%
%

%
%



\section{Introduction}
There are many ways humans observe the dynamic change in environmental phenomena of the real world, even though it is mostly short seasonal forecast. We may argue that it is effective to a certain degree. However, accurately forecasting a dynamic environmental event such as drought will involve the use of interconnected remote devices such as sensors to measure the environmental parameters. Technological advancement in Wireless Sensor Networks (WSN) has facilitated its use in environmental monitoring, habitat monitoring and tsunami warning systems \cite{chong2003sensor}. WSNs are networks of interconnected sensors that monitor environmental phenomena in geographic space irrespective of the topographical location. They have become an invaluable component of realizing an IoT-based environmental monitoring system; they form the 'digital skin' through which to 'sense' and collect the context of the surroundings and provides information on the process leading to events such as drought and weather changes.  However, these environmental properties are measured by various heterogeneous sensors of different modalities in distributed locations making up the WSN, using different terms in most cases to denote the same observed property \cite{kuhn2005geospatial}, \cite{akanbi2014semantic}, \cite{devarajutowards}. Also, information communities often use abstruse terms and vocabulary to categorize events \cite{llaves2014event}. These causes data heterogeneity problem, classified into \textit{naming heterogeneity} and \textit{cognitive heterogeneity} \cite{kuhn2009functional}. For example, water level property name is 'Hoehe' (in German) or 'Stav' (in Czech).  This is a major challenge hampering the realization of WSN solutions for environmental monitoring and consequently IoT, and causes lack of seamless data sharing and full integration of interconnected heterogeneous ubiquitous devices.

In recent years, scientists have started to investigate how to forecast a drought event accurately. This is necessary in order to mitigate the disastrous effect of drought in a particular geographic area. However, environmental phenomena such as droughts are complex. They evolve over space and time. According to \cite{peuquet1995event}, the greatest challenge is designing a framework which can track information about the 'what', 'where' and 'when' of environmental phenomena and the representation of the various dynamic aspects of the phenomena. The representation of such phenomena requires better understanding of the 'process' that leads to the 'event'. For example, a \textit {soil moisture sensor} provides sets of values for the observed property \textit {soil moisture}. The measured property can also be influenced by the \textit {temperature heat index } measured over the observed period. This makes accurate prediction based on these sensor values almost impossible without understanding the semantics and relationships that exist between these various properties, because various processes can lead to an environmental event such as drought. Moreover, research \cite{mugabe2010use}, \cite{masinde2011itiki} on indigenous knowledge (IK) on droughts has pointed to the fact that IK, e.g., on worms like  \textit {sifennefene worms} and plants like the \textit{mutiga tree}\footnote{A tree indigenous to South Africa} can indicate drier or wetter conditions, which can imply the likely occurrence of drought event over time \cite{sillitoe1998development}. This scenario shows that environmental events can be inferred from sensors' data, if proper semantic meaning is attached to it and augmented with some set of indicators derived from the IK.

Considering the aforementioned, it can be concluded that the key to improve the accuracy of forecasting a drought event is the understanding of 'space-time' interactions of variables with processes, ontology representation of the domain and semantic integration of the heterogeneous sensor data with indigenous knowledge for efficient drought forecasting. In order to provide services from heterogeneous data sources, it is necessary to build systems that attach meaning (semantics) to the sensors data. Therefore, an ontology-based semantic middleware is required to semantically represents the heterogeneous sensors data in machine-readable language and a reasoning engine that will infer patterns leading to drought event based on IK for an efficient drought early warning system (DEWS).

\section{MOTIVATION AND PROBLEM STATEMENTS}
This research was motivated by the following factors:
\\
\\
\textit {Lack of ontology-based semantic middleware for the representation of environmental process.}
\\
\\
According to \cite{noy2004semantic}, ontology is the formal description of a domain of discourse, intended for sharing among different applications, and expressed in a language that can be used for reasoning. Ontological modelling of key concepts of environmental phenomena such as \textit{object}, \textit{state}, \textit{process} and \textit{event}, ensures the drawing of accurate inference from the sequence of processes that lead to an event. This can easily be achieved by using an upper-level ontology that command wide-spread acceptance from the semantic community and extending it to meet the current environmental domain requirements. This ensures the development of high quality environmental ontology with well-defined vocabularies that allow explicit representation of the process, events and also attach semantics to the participants in the environmental domain. However, the lack of effective descriptive environmental ontology still persists, which gives rise to the problem of incompatibility, non-interoperability, lack of service integration, and difficulty in generating environmental inference from sensory readings since they are just raw data. Hence, there is a need for semantic middleware that attaches detailed semantic meaning to the raw data for ease of communication and knowledge sharing. Existing work on this can be found in \cite{eisenhauer2010hydra}, \cite{valls2013iland}, \cite{botts2008ogc}.
\\
\\
\textit {Semantic integration of various heterogeneous data sources with indigenous knowledge for an accurate environment forecasting.}
\\
\\
Studies reveal that over 80\%  of farmers in some parts of Kenya, Zambia, Zimbabwe and South Africa rely on Indigenous knowledge forecasts (IKF) for their agricultural practices \cite{masinde2011itiki}. IFKs provides an uncertain level of accuracy that could result in loss of yield and manpower. An IoT-based environmental monitoring system made up of interconnected heterogeneous weather information sources such as sensors, mobile phones, conventional weather stations, indigenous knowledge could alleviate this \cite{masinde2012role}. Large number of sensors/things could provides environmental data streams required to be semantically represented for seamless data integration with existing indigenous knowledge. This integration will improve the accuracy of predicting drought.
\\
\\
\textit {IoT-based forecasts communication and dissemination.}
\\
\\
There is a lack of effective dissemination channels for drought forecasting information, for example, the absence of smart billboards placed at strategic locations, smart phones, IP radios and semantic web. The output channels would ensure farmers have access to drought forecasting information and the spatial distribution of \textit{drought vulnerability index} would be effectively disseminated. However, the lack of this is a formidable challenge for an effective IoT-based environmental monitoring system.

\section{RESEARCH QUESTIONS}
The research will be developed based on the following questions:
\\
\\
\textit {To what extent does the adoption of knowledge representation and semantic technology in the development of a middleware enable seamless sharing and exchange of data among heterogeneous IoT entities?}
\\
\\
In recent years, the amount of computerized data and information available on the Web has spiraled out of control. Many different models and formats are being used that are incompatible with each other. Traditionally, the easiest way to address interoperability is to define standards [14]. Several standards have been created to cope with the data heterogeneities. Examples are the Sensor Markup Language (SensorML)\footnote{http://www.opengeospatial.org/standards/om} and Observations and Measurements Encoding Standard\footnote{http://www.opengeospatial.org/sensorml}, WaterML\footnote{http://www.his.cuahsi.org/wofws.html}, and American Federal Geographic Data (FGDC) Standard\footnote{https://www.fgdc.gov/metadata}. However, these standards provides sensor data to a predefined application in a standardized format only, and hence do not generally solve data heterogeneity. The promising technology to tackle these problems of heterogeneity and integration of ubiquitous data sources is semantic technologies. Semantic technologies have a stronger approach to interoperability than contemporary standards-based approaches \cite{oberle2004semantic}. It creates knowledge representation models that are general in order to allow meaningful information exchange among machines through detailed semantic referencing of metadata. It utilizes machine-readable languages such as Resource Description Framework (RDF) and Ontology Web Language (OWL) for seamless data sharing and integration in an event-driven way. Several upper-level ontologies are available for designating the concepts of object, process and events \cite{noy2004semantic}. Analyzing with top-level ontology is however very necessary to identify the basic objects and process that leads to an event. 
\\
\\
\textit {What are the main components of an implementation framework/architecture that employs the middleware to implement an IoT-based Drought Early Warning Systems (DEWS)?}
\\
\\
To answer this question, a case study based on environmental monitoring and drought forecasting is identified. Here, we examine the existing components and architecture used in the case study as it conforms to IoT based systems: The existence of ontology with well-defined vocabularies that allows an explicit representation of process, events, and their participants to deal with environmental phenomena; the availability of a reasoning engine that generates inference based on input parameters; the representation and integration of the inputs in machine-readable formats. This is the main focus of the research - the development of an ontology-based semantic middleware for the representation of heterogeneous data sources and integration with IK for accurate drought forecasting.
\\
\\
\textit {What method is best suitable to predict drought event by combining heterogeneous sensor data with human indigenous knowledge for an accurate drought forecasting system?}
\\
\\
Currently, most drought predicting/forecasting system is based on statistical model using data from weather stations and WSNs data only. This research potential of integrating semantically represented heterogeneous sensors data with IK is exciting, and will improve the overall accuracy of drought forecasting system.

\section{HIGH-LEVEL APPROACH TO MIDDLEWARE DESIGN}

The development of the semantic middleware follows a step-wise approach. The proposed middleware is a software layer composed of a set of various sub-layers interposed between the application layer and the physical layer. The essential role of the middleware is to hide the complexities and eliminate the data heterogeneity from multiple data sources \cite{kostelnik2011semantic}, and representing the data semantically based on a unified ontology. It also provides Application Programming Interface (API) for physical layer communication, abstraction of complex network communication and presenting the data in a machine readable format for easy integration and interoperability. 

Based on these needs, an environmental process-based ontology is required to overcome the problems associated with the dynamic nature of environmental data and the data heterogeneities. The study proposes to use a upper-level foundational ontology DOLCE (Descriptive Ontology for Linguistic and Cognitive Engineering) \cite{masolo2002wonderweb}, for the modelling of the foundational entities needed to represent the dynamic phenomena. Figure 2 depicts the ontology library. The entities will be identified and classified based on DOLCE classification of endurants, perdurants and quality \cite{gangemi2005ontology}. Semantically represented information from the sensor data streams will be integrated with IK using Complex Events Processing \textit{(CEP)} Engine \footnote{https://en.wikipedia.org/wiki/Complex\_event\_processing}. This will serve as the reasoning engine for inferring patterns leading to drought event, based on a set of rules derived from IK of the local people on drought. The measured properties are modelled by DOLCE to attach semantic meaning to the observed properties. The properties in data streams are fed into the \textit{CEP engine} to infer patterns. The semantic reasoning capability of the \textit{CEP engine} determines the patterns leading to the drought event. The information in form of \textit{drought vulnerability index} is disseminated to the targeted end-user via various output IoT channels such as the smart screen, semantic web and mobile apps. Finally, the overall system will be implemented and tested across a IoT-based environmental monitoring sensor network test bed. The domain of this particular case study is Free State Province, South Africa - an ongoing research project by AfriCRID \footnote{http://www.africrid.com/}(The Africa Research Unit for Research on Informatics for Drought), Department of Information Technology, Central University of Technology, Free State, South Africa.
 
\begin{figure}[ht!] 
\centering
\includegraphics[scale=0.5]{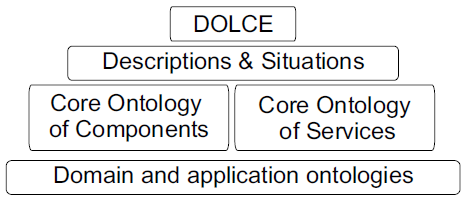}
\caption{The Ontology Library.}
\end{figure} 

\subsection{Middleware Development}
According to \cite{hornsby2008linking}, an ontology is a desirable solution for achieving semantic interoperability since it captures concepts of a domain and provides a foundation for discovering and resolving semantic conflicts in the sensor data. It consist of formal axioms that describe concepts, individuals and the relationships between them. The development of a middleware framework will be based on ontology. By using ontologies to enrich the descriptions of the domain, the semantics of data content or service functionality become machine-interpretable, and users are enabled to pose concise and expressive queries. This implies that the ontology developed will identifies entities and physical properties, followed by the relations between them for the semantic referencing of the metadata. Since, ontology is merely a passive object, an inference engine is also required to enable querying and reasoning with the semantic descriptions of sensor devices and services. A \textit{CEP engine}, which basically infer patterns based on a set of syntactic derivation rules from indigenous knowledge is used. An integrated development environment (IDE), e.g. Eclipse, JBuilder, allows the querying of the ontology infrastructure residing in the middleware layer. The resulted information is passed on to the output channels. Figure 2 depicts the middleware integration framework.
\\
\begin{figure}[ht!] 
\centering
\includegraphics[scale=0.43]{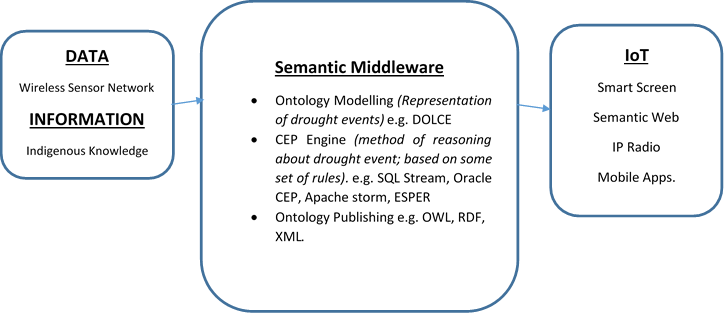}
\caption{The semantic middleware integration framework.}
\end{figure}
\subsection{Middleware Architecture}
The middleware layer is a three-tier architecture consisting of \textit{application abstraction layer}, \textit{ontology segment layer}, and the \textit{interface protocol layer} as shown in figure 3.
\subsubsection{Application Abstraction Layer}
This layer provides a high level of software abstraction that allows communication among the applications and the semantic middleware.
\subsubsection{Ontology Segment Layer}
The middleware layer semantically reference the heterogeneous data streams based on the unified ontology. This layer contains the ontology module, reasoning module, inference engine, and semantic services description module.
\subsubsection{Interface Protocol Layer}
The Interface protocols liaise with the storage database in the cloud for downloading the semi-processed sensory reading.
\begin{figure}[ht!] 
\centering
\includegraphics[scale=0.5]{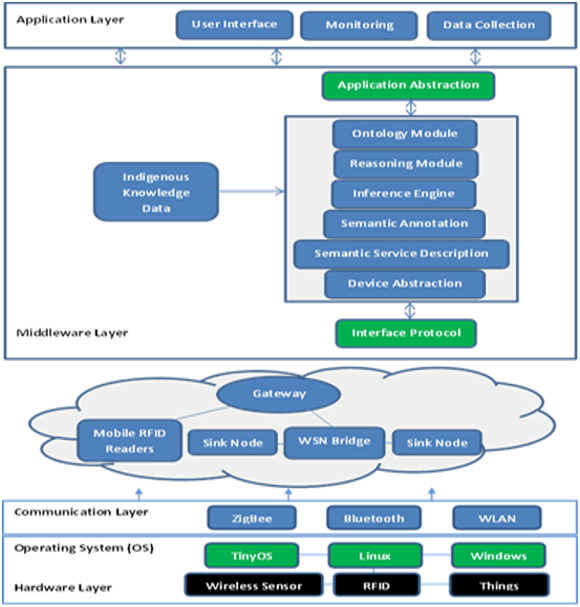}
\caption{Architecture of the middlware layer.}
\end{figure}

\section{OUTLOOK ON IMPLEMENTATION}
The first step involves data gathering from the sensors. This involves setting up WSNs for the area under currently under study. The research is using Libelium Sensor nodes especially the Waspmote Mote Runner\footnote{http://www.libelium.com/}. The motes are based on \textit{6LoWPAN} protocol which enables transmission of compressed IPv6 packets over IEEE 802.15.4 networks. The environmental readings are uploaded via SMS gateway for storage in the cloud. This will be followed by gathering the indigenous knowledge of the local people about drought, through the use of questionnaire, workshop and interactive sessions. The middleware layer acts as a bond joining heterogeneous domains of application community over heterogeneous interfaces. It incorporates interface protocols, which liaise with the storage database in the cloud for downloading the semi-processed sensory reading to be semantically represented based on the ontology through a mediator device \cite{christian2013event}. The semantically represented heterogeneous sensor data is integrated with the IK using the \textit{CEP engine}. The \textit{CEP engine} infer patterns leading to drought event based on the set of rules derived from the IK of the local people on drought. This would be used to establish accurate drought development patterns as early as possible and provide sufficient information to decision-makers to prepare for the droughts long before they happen. This way, the prediction can be used to mitigate effects of droughts. 

\section{EXPECTED OUTCOMES}
The study is expected to produce an ontology-based semantic middleware that will facilitate the semantic representation and integration of heterogeneous sensor data with indigenous knowledge. It will enhance effective integration in a heterogeneous environment, and most especially for the seamless data sharing and communication for an IoT-based DEWS. The middleware will be used to attach semantic meaning to observed property and the \textit{detection-oriented CEP Engine} will infer patterns from the observed properties based on set of a indicators derived from the indigenous knowledge. This will increase the accuracy of drought forecasting information that is disseminated via output channels. Also, the representation of the heterogeneous data sources by the proposed ontology and integration with indigenous knowledge will be demonstrated.

\section{ACKNOWLEDGMENT}
The PhD research is funded by the Research Grant Scheme of Central University of Technology, Free State, South Africa. We also wish to thank anonymous reviewers for their valuable comments on earlier versions of this paper.

%
\bibliographystyle{abbrv}
\bibliography{sigproc}  
%
%

\end{document}